\documentclass[10pt,twocolumn,letterpaper]{article}

\pdfoutput=1

\usepackage{cvpr}
\usepackage{times}
\usepackage{epsfig}
\usepackage{graphicx}
\usepackage{amsmath}
\usepackage{amssymb}
\usepackage{color}
\usepackage{nccmath}
\usepackage{textcomp}

\usepackage[breaklinks=true,bookmarks=false]{hyperref}

\cvprfinalcopy %

\input{macros}

\DeclareMathOperator*{\argmin}{arg\,min}

\begin{document}

\title{Monocular Reconstruction of 
Neural Face Reflectance Fields}

\author{ Mallikarjun B R$^{1}$ \quad Ayush Tewari$^{1}$ \quad Tae-Hyun Oh$^{2}$ \quad Tim Weyrich$^{3}$ \\
\quad Bernd Bickel $^{4}$ \quad Hans-Peter Seidel $^{1}$ \quad Hanspeter Pfister $^{5}$ \\
\quad Wojciech Matusik $^{6}$ \quad Mohamed Elgharib$^{1}$ \quad Christian Theobalt$^{1}$ \\
\\
$^1$ Max Planck Institute for Informatics, Saarland Informatics Campus \quad 
$^2$ POSTECH \quad \\
$^3$ University College London\quad 
$^4$ IST Austria \quad 
$^5$ Harvard University\quad 
$^6$ MIT CSAIL\\
}

\newcommand{\myparagraph}[1]{\vspace{-2pt} \noindent \textbf{#1}}

\maketitle

\begin{abstract}
The reflectance field of a face describes the reflectance properties responsible for complex lighting effects including diffuse, specular, inter-reflection and self shadowing. 
Most existing methods for estimating the face reflectance from a monocular image assume faces to be diffuse with very few approaches adding a specular component. 
This still leaves out important perceptual aspects of reflectance as higher-order global illumination effects and self-shadowing are not modeled. 
We present a new neural representation for face reflectance where we can estimate all components of the reflectance responsible for the final appearance from a single monocular image. 
Instead of modeling each component of the reflectance separately using parametric models, our neural representation allows us to generate a basis set of faces in a geometric deformation-invariant space, parameterized by the input light direction, viewpoint and face geometry. 
We learn to reconstruct this reflectance field of a face just from a monocular image, which can be used to render the face from any viewpoint in any light condition.
Our method is trained on a light-stage training dataset, which captures 300 people illuminated with 150 light conditions from 8 viewpoints. 
We show that our method outperforms existing monocular reflectance reconstruction methods, in terms of photorealism due to 
better capturing of physical premitives, such as sub-surface scattering, specularities, self-shadows and other higher-order effects. 
\end{abstract}

\section{Introduction}
Monocular face reconstruction (i.e. dense reconstruction of 3D face geometry, reflectance and illumination) 
has vast applications in visual effects, telepresence, portrait relighting, facial reenactment and interactions in virtual environments.
It has been an active area of research with tremendous progress in all aspects of reconstruction, including both geometry and reflectance. 
Our focus is on the reconstruction of the face reflectance, which captures the interaction between the face and scene illumination, playing a very important role in perception.
In the literature, one category of methods \cite{GZCVVPT16,tewari17MoFA,Tran_2017_CVPR}, approximates faces as a Lambertian surface. 
Many of them use analysis-by-synthesis optimization to estimate face geometry, spherical harmonics lighting, and diffuse face reflectance; the latter is a stark simplification of true face reflectance.
This type of representation fails to capture important specularities and sub-surface effects in face reflectance, which prevents truly photorealistic reconstruction. 
While some approaches~\cite{Schneider17,aldrian2012inverse} use ambient occlusion and precomputed radiance transfer to model shadows in an inverse rendering framework, they still assume simple reflectance properties of the face, which limit photorealism.
Another category of methods~\cite{yamaguchi_high-fidelity_2018,lattas2020avatarme} reconstruct diffuse and a specular face albedo from an image using machine learning methods.
While being more complete, this still leaves out important components of the reflectance, such as self shadowing and other higher-order view-dependent effects and sub-surface effects.

We present the first monocular face reconstruction algorithm that estimates a full face reflectance field, representing both \emph{view direction}- and \emph{light direction}-dependent reflectance properties, from a single face image. 
We train a CNN that infers the face reflectance field from a single image, and represents it as a basis set of images showing the illuminated face in a normalized space. 
The images, and thus the reflectance field, are parameterized by light direction, view direction and face geometry.
This is similar to the representations used by image-based techniques for acquiring reflectance fields~\cite{debevec2000acquiring,Meka19,Sun19,fyffe2009cosine}.
However, the crucial difference to our work is that they only capture light-dependent, not view-dependent effects; they can only relight the given input camera view.  
While~\cite{debevec2000acquiring} can render the face from a different viewpoint, doing so requires an assumption of the BRDF model of the face, and ignores effects such as self-shadowing in the reflectance. 
Our method goes significantly further by estimating
the full reflectance field, including view-dependent effects. 
We can change both the light source and viewpoint in the image. 
We do this by also jointly estimating
the 3D face geometry from the monocular image, and representing the basis images in the UV space~\cite{botsch2010polygon} of the template face mesh.
This also offers other advantages, such as generalization outside of the training data space. 
Our method is trained on a light-stage dataset, which captures 300~people illuminated with 150~point light sources one at a time, and from 8~viewpoints. 
All faces in the dataset are in a neutral expression with mouth closed. 
Our method still 
generalizes to real images with general facial 
expression, since the training is done in the normalized expression-invariant UV space. 

In summary we make the following contributions: 
\begin{itemize}
    \item A monocular method for estimating a deep face reflectance field. Our method is trained with a large set of light-stage data.  Reconstructed faces model complex pose/view- and scene illumination dependent appearance, beyond diffuse and specular reflectance.
    \item A new deep representation for face reflectance fields, allowing us to generalize to real-world images after training on a light stage dataset.
    This generalization is obtained by virtue of the explicit use of a canonical space invariant to pose, identity and expression, i.e., UV space, as well as training with data synthesized by natural environment maps.
\end{itemize}

\section{Related Work}
The literature on face reflectance capture is vast, with methods varying from requiring multi-view multi-illumination images as input~\cite{Meka19,debevec2000acquiring,mvfc_ghosh} to methods which can reconstruct reflectance from a single image. 
We focus our discussion on monocular methods.

\paragraph{Analysis-based Synthesis}
Many methods reconstruct face reflectance by solving an analysis-by-synthesis optimization problem minimizing the difference between model and single input image. 
Since this is an under-constrained problem, methods often make simplifying assumptions, such as the skin having Lambertian reflectance~\cite{tewari17MoFA,GarridoZWBPBT16,Tran_2017_CVPR,Tran18,Richardson_2017_CVPR}.
This allows them to represent lighting using coarse spherical harmonic illumination~\cite{Ramamoorthi2001}.
Some other methods use a Phong-reflectance assumption~\cite{Blanz1999,LWSLVDT13}, which can also model specularities.
Specularities using spherical harmonics have also been explored~\cite{aldrian2012inverse,smith2020morphable}.
These representations do not model effects such as sub-surface scattering and self-shadowing, which are important for face appearance.  
Some methods model shadows using precomputed radiance transfer~\cite{Schneider17} or ambient occlusions~\cite{aldrian2012inverse}. 
However, due to a Lambertian or simple specular assumption, the final output lacks photo-realism. 
Please refer to a recent survey~\cite{egger20193d} for more details on these methods. 

\paragraph{Supervised Learning}
Another class of methods are based on supervised learning.
Here, the training data is well-defined %
, captured from light stages featuring a dome of controlled lighting. 
At test time, the methods can reconstruct %
rich reflectance from monocular images.
The common representation here is to separate the reflectance into diffuse and specular albedo~\cite{yamaguchi_high-fidelity_2018,lattas2020avatarme}.
In Yamaguchi~\etal~\cite{yamaguchi_high-fidelity_2018} the solution also infers a high-frequency displacement map representing mesoscopic surface details.
In Lattas~\etal~\cite{lattas2020avatarme} separate networks estimate the specular albedo and normals from the diffuse albedo and the 3DMM shape normals.
However, other complex effects such as self shadows and view-dependent inter-reflectance cannot be captured. 
A computationally expensive step of path tracing is performed to simulate shadows at test time. 

\paragraph{Image-based Methods}
Here we review supervised methods that train either using light stage training data~\cite{Sun19,Meka19} or just on monocular images~\cite{Zhou_2019_ICCV}. 
Meka \etal~\cite{Meka19} show that two spherical color gradient images are capable of modeling a 4D reflectance field. This includes high-frequency details and specular reflections. 
Sun \etal~\cite{Sun19} present an encoder-decoder architecture for manipulating the lighting of an input. The network is trained on light-stage data of 18 subjects captured under several light sources. 
Zhou~\etal~\cite{Zhou_2019_ICCV} utilize a spherical harmonic representation of lighting to synthesize large-scale training data for a relighting network. 
While image-based approaches provide a range of capabilities, they directly work on input images and not in 3D.
This does not allow to capture the full reflectance field; these methods can only relight a given image, but not change the viewpoint.

Our method, on the other hand, allows us to reconstruct the full reflectance field from a monocular image, thus allowing control over both light and viewpoint. 
We do not make any assumptions about the reflectance properties of the face, and can thus capture all effects including sub-surface scattering, specularities and self shadows.

\begin{figure*}
\begin{center}
   \includegraphics[width=\textwidth]{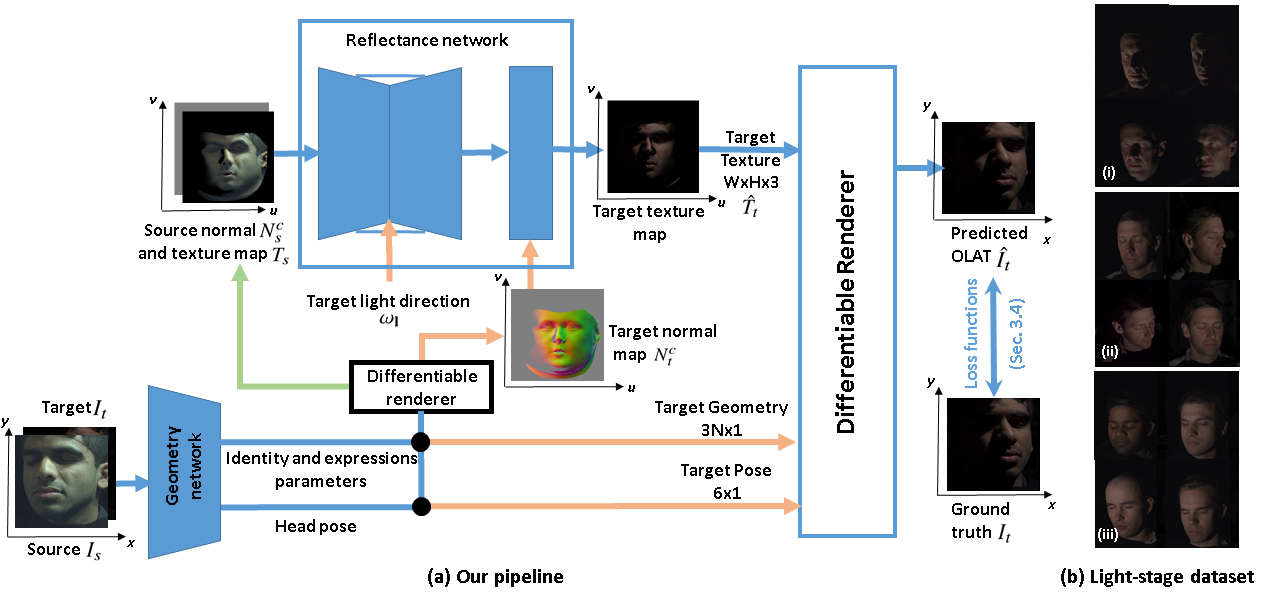}
\end{center}
\caption{
(a) Our approach learns the full face reflectance field by reconstructing an input image with different head poses and point-source lightnings (see predicted OLAT). At inference this allows us to synthesize results with any environment map by linearly combining different OLAT predictions. Our solution is formulated within a normalized UV-space and minimizes for several loss functions through a differentiable renderer. Note the geometry network processes both the source and target images.
(b) Our solution is trained with a light-stage dataset which includes 150 lighting conditions (i), with 8 camera-views (ii) and 350 subjects (iii). We use 300 subjects for training, 10 for validation and the rest for test.
}
\label{fig:pipeline}
\end{figure*}

\section{Method}
Our method takes as input an in-the-wild image of a face, a target point light source direction and the target viewpoint. The output of the network is a mesh of the face lit by a point light from the desired direction which can be rendered from the target viewpoint. 
At test time, we can render the reconstructed face geometry from any viewpoint and under any environment map by projecting the environment map on a densely sampled point light basis. 
\subsection{Dataset}
\label{sec:dataset}

Our data-driven approach learns to predict the face reflectance field, which is a function of the face geometry, light sources and camera pose.
We train our model on a light-stage dataset~\cite{Weyrich2006Analysis} consisting of HDR images of 350 identities, captured with 8 cameras distributed in front of the face on a hemisphere (see Fig.~\ref{fig:pipeline}-b). 
The light stage also contains 150 point light sources uniformly placed on the sphere surrounding the face. 
150 images are captured per person and per camera, with each of the light sources turned on one light at a time (so-called OLAT images). 
Every subject was captured with neutral expression with eyes and mouth closed.
In order to simulate data that look like in-the-wild images under natural illumination,
we relight the light stage data using HDR environment maps.
In particular we use a combination of around 205 Laval Outdoor~\cite{hold2019deep} and around 2233 Laval Indoor HDR~\cite{gardner2017learning} images, as done in~\cite{Sun19}.
Our training dataset includes 1000 relit images each, for 300 identities. For each of the relit images, we have a randomly selected OLAT from a random camera view as target image.
We use images of $10$ identities for validation, and the rest $40$ identities for test. 
Our reflectance field representation operates in a normalised UV space for facial geometry. This enables generalization of our approach to arbitrary face expressions, despite all training data showing neutral face expressions. 

\subsection{Reflectance Field Representation}
\label{sec:reflectance_field_repr}
Our reflectance field is a function $\mathcal{R}(\mathcal{G}, \mathbf{\omega_v}, \mathbf{\omega_l})$, describing the reflectance of a face with geometry $\mathcal{G}$, under viewing direction $\mathbf{\omega_v}$ and illuminated by an input point light source direction $\mathbf{\omega_l}$, where $\mathbf{\omega_v}$ and $\mathbf{\omega_l}$ are unit norm vectors.
We represent the face geometry using a 3D Morphable Model~\cite{Blanz1999}, which includes an identity model $M_\text{id} \in \mathbb{R}^{3N \times m_i}$ and an expression model $M_\text{exp} \in \mathbb{R}^{3N \times m_e}$, where $N$ is the number of vertices. 
The vectors of $M_\text{id}$ and $M_\text{exp}$ are scaled with their corresponding standard deviations, as in~\cite{tewari17MoFA}.
This representation is well-suited for monocular reconstruction~\cite{tewari17MoFA,Tewari19FML,Tran19}.
Mesh vertices are represented by $\mathbf{v}$, $|\mathbf{v}| = 3N$.
The final geometry is defined as
\begin{align}
  \mathbf{v} (\alpha, \beta ; M_\text{id}, M_\text{exp}) &=  \bar{v} +  M_\text{id} \alpha + M_\text{exp} \beta
  \enspace{.}\nonumber
\end{align}
We use the mean mesh $ \bar{v}$ from~\cite{Blanz1999}; $\alpha \in \mathbb{R}^{m_i}$ and $\beta \in \mathbb{R}^{m_e}$ are the identity and expression parameters. 
In monocular reconstruction, it is not possible to separate the effects of head and camera pose. 
We remove this ambiguity by assuming a camera with fixed extrinsics and intrinsics, and only modeling head pose $\mathbf{\omega_h} \in \text{SO}(3)$ as variable.
Although the reflectance does not depend on the global translation, we need it to render the face in the correct position in the image. 
For any vertex $\mathbf{v_i} \in \mathbb{R}^3$, we can compute the camera space coordinates $\mathbf{v_i^c} = \mathbf{\omega_h} \mathbf{v_i} + \mathbf{t}$, where $\mathbf{t} \in \mathbb{R}^3$ is the global translation. 
The complete geometry can be represented as $\mathbf{v^c} \in \mathbb{R}^{3N}$, with 
$\mathbf{v_i^c}$, $\forall i \in \{0,{\cdots},N\}$ 
stacked together.
The reflectance field can then be represented as $\mathcal{R}(\mathbf{v^c}, \mathbf{\omega_l})$.
We represent the output of this function 
as a $512 \times 512$ RGB image in a normalized UV parametrized space, defined using the template mesh used to represent $\mathbf{v}$, see Fig.~\ref{fig:pipeline}-a.
This is a pose, expression and identity deformation-invariant representation, allowing us to easily generalize to in-the-wild images of varying identity and expression.
In addition, it allows us to use a U-Net architecture~\cite{ronneberger2015u}, since the pixel correspondences required for the skip connections are valid irrespective of target head pose.

\subsection{Network Architecture}
Our framework consists of two neural networks, the \emph{Geometry Network} and the \emph{Reflectance Network}, as shown in Fig.~\ref{fig:pipeline}-a. 
Each sample in our training consists of two images, source ($I_s$) and target ($I_t$).
$I_s$ is an image lit by a natural environment map and $I_t$ is the image of the same person in the same or different pose, under one of the 150 different OLAT lighting condition.

The \emph{Geometry Network} takes both source and target face images as input and reconstructs the 3D face geometry, represented as pose, identity and expression parameters of the 3DMM. 
Given the reconstructed face geometry of the source image in camera-space coordinates, a differentiable renderer produces a source texture map $T_s \in \mathbb{R}^{512\times512}$ in the UV space. 
Our goal is to generate an OLAT image in the UV space, lit from a light source  with direction $\mathbf{\omega_l}$ and with head pose $\mathbf{\omega_h}$.
From the camera space geometries $\mathbf{v}_s^c$ and $\mathbf{v}_t^c$ of the source and target images, we also compute the source and target surface normal maps $N_s^c \in \mathbb{R}^{512\times512}$ and $N_t^c \in \mathbb{R}^{512\times512}$.
The \emph{Reflectance Network} takes as input $T_s$, $N_s^c$, $\mathbf{\omega_l}$ and $N_t^c$, as shown in Fig.~\ref{fig:pipeline}-a, and outputs the target texture map $\hat{T_t}$ in a normalized UV space i.e., every pixel corresponds to a semantically well-defined structure such as eye corner or nose. 
The network produces an OLAT texture as output, which is rendered using the target geometry and pose to compute the final rendererd image $\hat{I_t}$. 

The \emph{Geometry Network} is based on AlexNet~\cite{KrizhSH2012,tewari17MoFA}, while the \emph{Reflectance Network} is based on a U-Net architecture~\cite{Ronneberger15}. The U-Net consists of 8 down and up convolution layers with skip connections and kernels of spatial dimensions $3 \times 3$. 
This is followed by 5 convolutional layers with a stride 1, which takes the output features, as well as the target normal map as input (see Fig.~\ref{fig:pipeline}-a). 
Note that the target lighting is fed to the U-Net bottleneck.

Our differentiable renderer renders a 2D image from a 3D face mesh.
We estimate the visible triangles using a z-buffering algorithm. 
Texture mapping is used to compute the color values. 
Interpolation (both on the mesh and the texture map) is done using barycentric coordinates. 
The differentiable renderer offers means for backpropagating the gradients through our normalized representation and thus allows our loss functions to be defined in image space (Sec.~\ref{sec:losses}) 
Our differentiable renderer is implemented as a data-parallel custom TensorFlow layer.

\subsection{Loss Functions}
\label{sec:losses}
We enforce several loss functions to enable the learning of the face reflectance field. 
Our method concurrently learns to estimate the geometry and head pose as well. %
\begin{align}
\nonumber
    \mathcal{L}(I_s, I_t, \mathbf{\omega_l}, \mathbf{\theta_n}) =  \lambda_\text{l} \mathcal{L}_\text{l}(I_s, I_t, \mathbf{\theta_n})  + \lambda_\text{r} \mathcal{L}_\text{r}(I_s, I_t, \mathbf{\theta_n}) +  \\
    \lambda_\text{p} \mathcal{L}_\text{p}(I_s, I_t, \mathbf{\omega_l}, \mathbf{\theta_n}) + \lambda_\text{f} \mathcal{L}_\text{f}(I_s, I_t, \mathbf{\omega_l}, \mathbf{\theta_n})
    \enspace{.}
\end{align}
Here, $\mathbf{\theta_n}$ are the trainable network parameters for both geometry and reflectance networks, 
$\mathcal{L}_\text{l}$ is a landmark alignment term, $\mathcal{L}_\text{r}$ is a geometry regularization term,
$\mathcal{L}_\text{p}$ is a photometric alignment term and $\mathcal{L}_\text{f}$ is a deep feature alignment term. %

\paragraph{Landmark loss} This loss provides a strong geometric cue for the 3D geometry reconstruction task. 
\begin{align}
\nonumber
    \mathcal{L}_\text{l}(I_s, I_t, \mathbf{\theta_n}) =
    \|L(\mathbf{v_s^c}(I_s,\mathbf{\theta_n})) - L_s \|_2^2 + \\
    \|L(\mathbf{v_t^c}(I_t,\mathbf{\theta_n})) - L_t \|_2^2
    \enspace{.}
\end{align}
We use 66 automatically detected landmarks~\cite{SaragihLC11a} from the source and target images, $L_s$ and $L_t$ as the ground truth.
The landmarks from the reconstructions, $L(\mathbf{v_s^c})$ and $L(\mathbf{v_t^c})$ are computed by projecting the annotated landmarks on the mesh to the image plane using the fixed camera parameters. 
Contour landmarks cannot be fixed since they slide on the mesh, so we compute these landmarks as the closest mesh vertices from the estimated 2D landmarks~\cite{tewari2018self}.

\paragraph{Geometry Regularization} We use common regularizers used in monocular geometry reconstruction:
\begin{align}
\nonumber
    \mathcal{L}_\text{r}(I_s, I_t, \mathbf{\theta_n}) =
    \sum\nolimits_{i=\{s,t\}} \lambda_{\alpha} \| \alpha_i(I_i, \mathbf{\theta_n}) \|_2^2 + \\
    \lambda_{\beta} \| \beta_i(I_i, \mathbf{\theta_n}) \|_2^2 
    \enspace{.}
\end{align}
This loss ensures that the final geometry is plausible.

\begin{figure*}
\centering
   \includegraphics[width=0.9\textwidth]{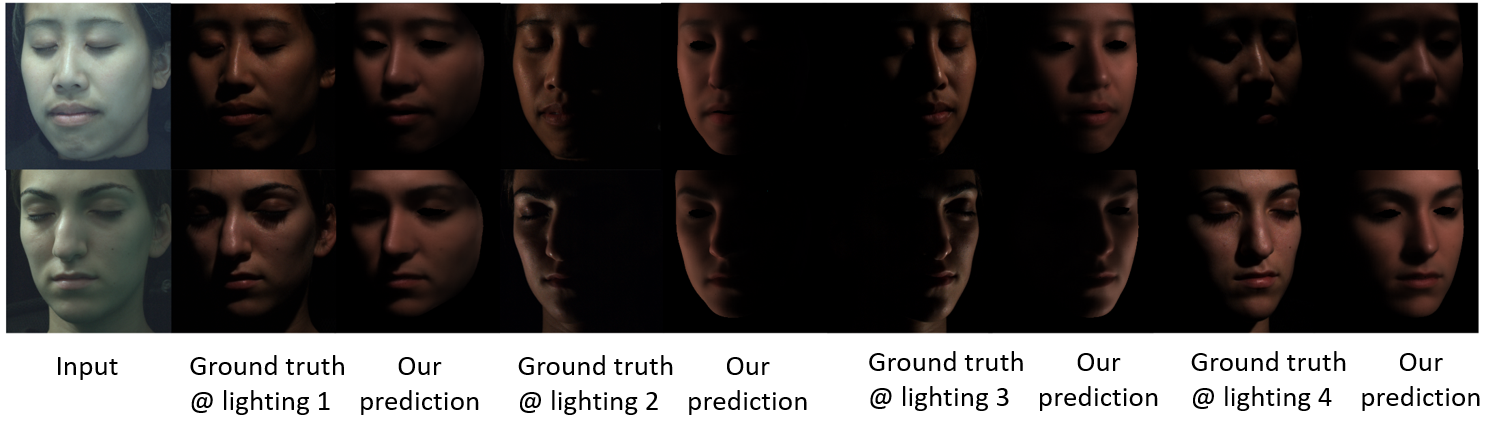}
\caption{Input image (left) and renderings under different point source lights and with different head poses. Our results resemble ground truth with accurate shadows. Input is taken from the light stage data-set where ground truth is available.
}
\label{fig:results1}
\end{figure*}

\begin{figure*}
\centering
   \includegraphics[width=0.9\textwidth]{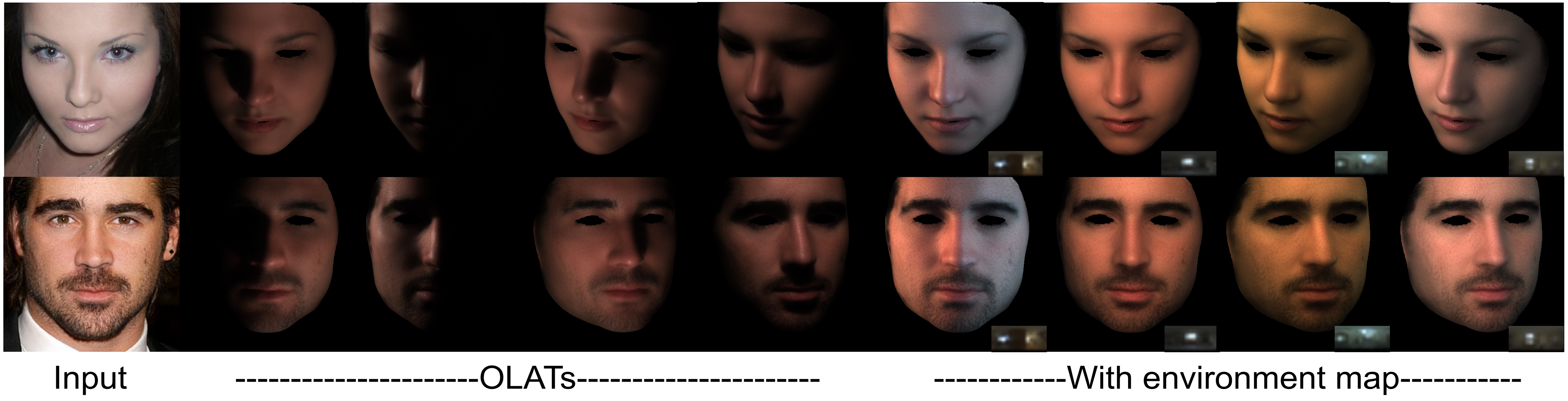}
\caption{Input image (left) and its OLATs with same pose (2nd and 3rd) and different pose (4th and 5th). Similarly, we have the input image relighted using random environment map (bottow right inset) with same pose (6th and 7th) and different pose(8th and 9th). The scene illumination is identical in each column , allowing us to observe the view dependent effects. For example, observe the the change is position of dominant specularity spot on the nose in column 4.}
\label{fig:results2}
\end{figure*}

\paragraph{Photometric loss} This loss ensures that the final relit images are close to the ground truth.
\begin{equation}
    \mathcal{L}_\text{p}(I_s, I_t, \mathbf{\omega_l}, \mathbf{\theta_n}) =
    \|M_t(\mathbf{P}) \odot (\hat{I_t}(\mathbf{P}) - I_t) \|_1 
    \enspace{.}
\end{equation}
As explained earlier, the final rendered image $\hat{I_t}$ is parametrized using the source texture map $T_s$, the normal maps $N_s^c$ and $N_t^c$, and the light direction $\mathbf{\omega_l}$.
Thus, $\mathbf{P} = (T_s(I_s, \mathbf{\theta_n}), N_s^c(I_s, \mathbf{\theta_n}), N_t^c(I_t, \mathbf{\theta_n}), \mathbf{\omega_l})$
We only evaluate the loss in a masked interior face region $M_t(\mathbf{\omega_h}(I_t))$, computed using the renderer. 
$\odot$ is an element-wise multiplication operator.
The supervision for our UV space reflectance field is thus indirect through the final rendered image using differentiable rasterization.

\paragraph{Feature loss} The $\ell_1$ loss is known to oversmooth details~\cite{isola2017image}. 
To preserve the high-frequency details in the output, we introduce a deep feature loss~\cite{johnson2016perceptual} with two terms.
\begin{equation}
    \mathcal{L}_\text{f}(I_s, I_t, \mathbf{\omega_l}, \mathbf{\theta_n}) = \mathcal{L}_\text{I}(I_s, I_t, \mathbf{\omega_l}, \mathbf{\theta_n})  +  \mathcal{L}_\text{L}(I_s, I_t, \mathbf{\omega_l}, \mathbf{\theta_n})
    \enspace{.}
\end{equation}

To extract features and compute $\mathcal{L}_\text{I}$, we use the layers $F{=}\{\texttt{conv1\_2{,}{conv2\_2}{,}{conv3\_3}}\}$ of a VGG network $V_f$ pretrained on ImageNet~\cite{johnson2016perceptual} to constrain the output texture map and image as follows:
\begin{fleqn}
\begin{equation}
\nonumber
     \mathcal{L}_\text{I}(I_s, I_t, \mathbf{\omega_l}, \mathbf{\theta_n}) =  
\end{equation}
\begin{equation}
    \nonumber
    \sum\nolimits_{f\in{F}} \Big(\text{\small} \big\|V_f(M_t(\mathbf{P}) \odot \hat{I_t}(\mathbf{P})) - V_f(M_t(\mathbf{P}) \odot I_t) \big\|_2^2 
\end{equation}
\end{fleqn}
\begin{equation}
    + \big\|V_f(\hat{T_t}(\mathbf{P})) - V_f(T_t(I_t,\mathbf{\theta_n})) \big\|_2^2 \Big)
    \enspace{.}
\end{equation}

We use another feature loss from features of a VGG network $S_f$ trained to predict the light direction from images~\cite{Meka19}.
Specularities depend on light direction, thus the features learned for predicting the latter encode the necessary information:
\begin{equation}
    \mathcal{L}_\text{L}(I_s, I_t, \mathbf{\omega_l}, \mathbf{\theta_n})
    =
    \sum\nolimits_{f\in{F}} 
    \bigl\|S_f(\hat{T_t}(\mathbf{P})) - S_f(T_t(I_t,\mathbf{\theta_n})) \bigr\|_2^2 
    \enspace{.}
\end{equation}

\paragraph{Training} We minimize our loss function summed over all samples in the training dataset using mini-batch of size $1$ with Adadelta Optimizer~\cite{adaDeltaC} with a learning rate of $0.05$ in order to obtain the network weights $\mathbf{\theta_n}$.
We implement our method in Tensorflow~\cite{tensorflow2015-whitepaper}.
We set $\lambda_{\alpha}= 0.4$, $\lambda_{\beta}= 0.002$, $\lambda_\text{l} = 25$, $\lambda_\text{p} = 5$, $\lambda_\text{r}=1$ and $\lambda_\text{f}=1$.
To improve generalization of geometry reconstruction, we also include monocular images from FFHQ~\cite{karras2019style} in our training.
FFHQ is only used for
the geometry losses, $\mathcal{L}_\text{l}$ and $\mathcal{L}_\text{r}$, in this case. 
Overall $20\%$ of our batches are sampled from FFHQ, and the rest from the light-stage data. 
The \emph{reflectance network} is only trained on the light stage images.

\begin{figure*}
\centering
   \includegraphics[width=1.0\textwidth]{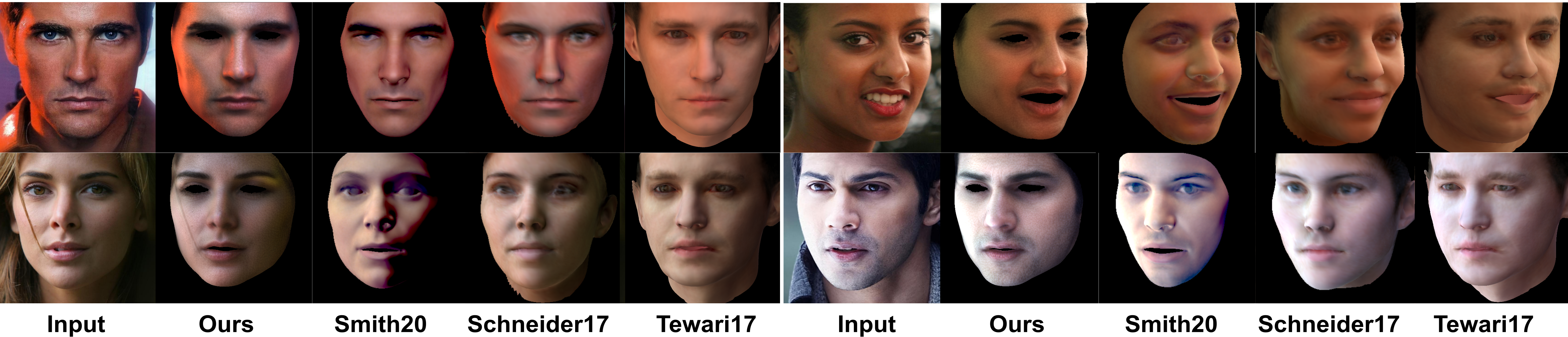}
\caption{Comparing our face reconstruction to the approaches of Smith~\etal~\cite{smith2020morphable}, Schneider~\etal~\cite{Schneider17} and Tewari~\etal~\cite{tewari17MoFA}. 
Our approach better captures specularities, sub-surface scattering, hard-shadows and overall produces more photorealistic results.
}
\label{fig:comp1}
\end{figure*}

\begin{table}[]
\centering
\begin{tabular}{|l|l|}
\hline
& \begin{tabular}[c]{@{}l@{}} Si-MSE (std. dev.)\end{tabular} 
\\ \hline
Same Pose                                 &  0.00070 ($\sigma$=0.00059)                                          \\ 
Different Pose        &             0.00084 ($\sigma$=0.00088)           \\ \hline
\end{tabular}
    \vspace{0.2cm}
	\caption{Reflectance reconstruction errors of our method, under the same and different head poses.
	}
\label{tab:olat_metric}	
\end{table}

\begin{figure*}
\centering
   \includegraphics[width=\textwidth]{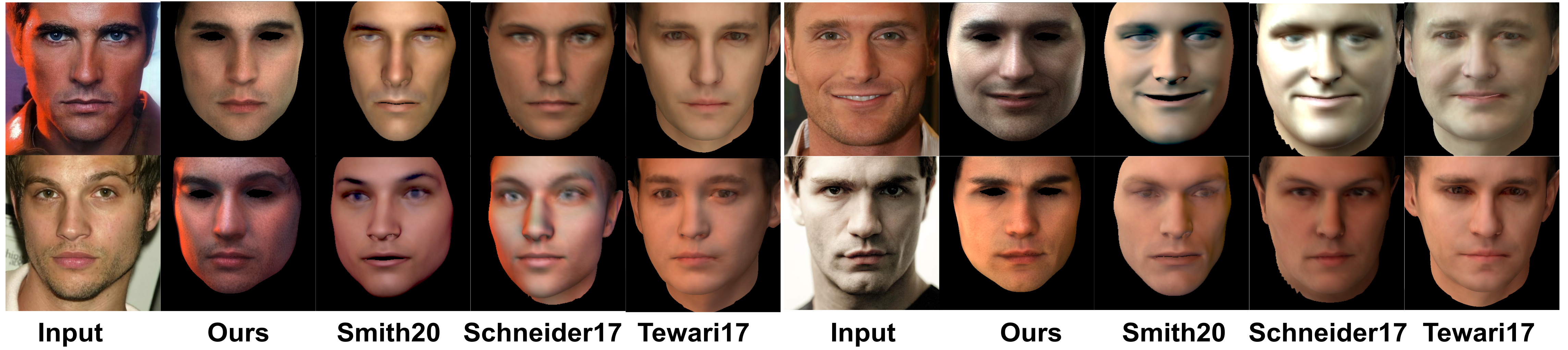}
\caption{Light transfer results between 2 different images. 
Each row shows the results of relighting the input image with the light estimated from the other row. 
Our approach relights an image and edits its head pose, all while maintaining its identity and facial integrity.
}
\label{fig:relight}
\end{figure*}

\begin{table*}[]
\centering
\resizebox{0.9\linewidth}{!}{
\begin{tabular}{|l|llll|}
\hline
& \begin{tabular}[c]{@{}l@{}} Ours \end{tabular} & \begin{tabular}[c]{@{}l@{}} Smith~\etal~\cite{smith2020morphable}\end{tabular} & \begin{tabular}[c]{@{}l@{}} Schneider~\etal~\cite{Schneider17} \end{tabular} & \begin{tabular}[c]{@{}l@{}} Tewari~\etal~\cite{tewari17MoFA}\end{tabular} \\ \hline
Reconstruction (Si-MSE)                                 &    \textbf{0.004} ($\sigma{=}${0.002})  & 0.017 ($\sigma{=}$0.015) & 0.010 ($\sigma{=}$0.008) & 0.007 ($\sigma{=}$0.003) \\ 
Transfer  (Si-MSE)                               &    \textbf{0.002} ($\sigma{=}${0.001}) &  0.018 ($\sigma{=}$0.011) & 0.007 ($\sigma{=}$0.005) & 0.003 ($\sigma{=}${0.001})  \\ 
Reconstruction (Face dis)                                 &    \textbf{0.550} ($\sigma{=}${0.080}) &  0.650 ($\sigma{=}$0.076) & 0.685 ($\sigma{=}$0.067) & 0.762 ($\sigma{=}${0.075})  \\ 
Transfer (Face dis)                                 &    \textbf{0.482} ($\sigma{=}${0.084}) & 0.605 ($\sigma{=}$0.092) & 0.644 ($\sigma{=}$0.077) & 0.689 ($\sigma{=}${0.076})  \\ 
\hline
\end{tabular}
}
	\caption{\label{tab:recon_metric} Reconstruction and reflectance transfer errors (in Si-MSE and Face dis with std. dev. $\sigma$) of our method, compared with the approaches of Smith~\etal~\cite{smith2020morphable}, Schneider~\etal~\cite{Schneider17} and Tewari~\etal~\cite{tewari17MoFA}.
	Evaluation is performed on 130 images from CelebA-HQ~\cite{CelebAHQ} for reconstruction, and on 86  images from our test set for reflectance transfer.
	}
\end{table*}

\begin{table*}[]
\centering
\begin{tabular}{|l|l|l|}
\hline & \begin{tabular}[c]{@{}l@{}} Without normal maps (std. dev.)\end{tabular}            & \begin{tabular}[c]{@{}l@{}} With normal maps (std. dev.)\end{tabular}   \\ \hline
Same Pose             &             0.00113 ($\sigma$=0.00093)                  &  \textbf{0.00070} ($\sigma$=0.00059)       \\ 
Different Pose        &             0.00126 ($\sigma$=0.00116)                  &  \textbf{0.00084} ($\sigma$=0.00088)       \\ \hline
\end{tabular}
	\caption{Reflectance reconstruction errors of our method, under the same and different input head poses. Removing the normal maps (source and target) from our network design clearly degrades performance.
	}
\label{tab:olat_metric_normal}	
\end{table*}

\begin{table*}[]
\centering
\begin{tabular}{|l|l|l|}
\hline & \begin{tabular}[c]{@{}l@{}} Only mean face (std. dev.)\end{tabular}            & \begin{tabular}[c]{@{}l@{}} With all components (std. dev.)\end{tabular}   \\ \hline
Reconstruction (Si-MSE)             &     0.011         ($\sigma$=0.005)                  &  \textbf{0.004} ($\sigma{=}${0.002})      \\ 
Reconstruction (Face dis)        &        0.550      ($\sigma$=0.073)                  &  \textbf{0.550} ($\sigma{=}${0.080})      \\ \hline
\end{tabular}
\vspace{0.1cm}
	\caption{
	Reflectance reconstruction errors of our method (in Si-MSE and Face dis with std. dev. $\sigma$) with and without face geometry learning. Performance degrades when only the mean face mesh is used (middle column), as opposed to learning the face geometry (last column).
	}
\label{tab:olat_metric_geom}	
\end{table*}

\begin{figure*}
\centering
   \includegraphics[width=0.9\textwidth]{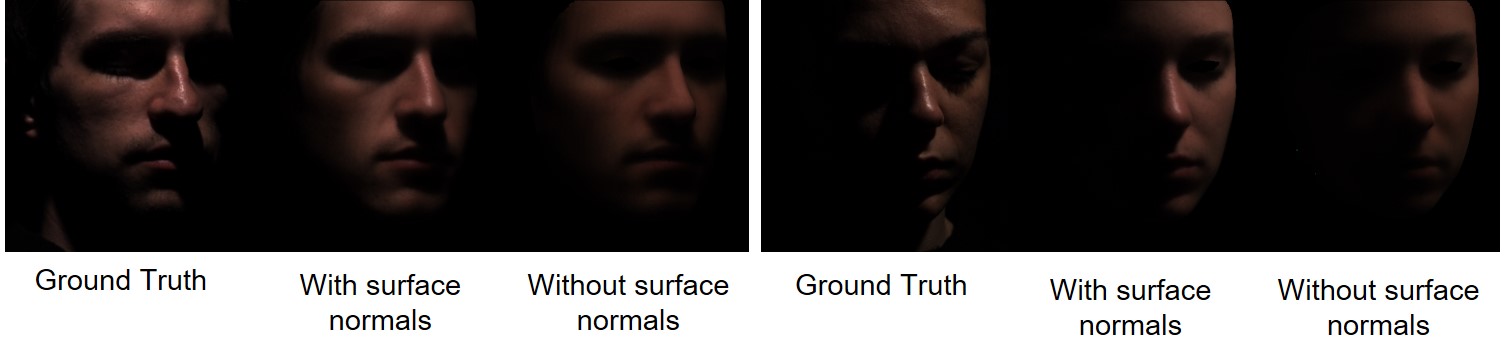}
\caption{Removing surface normals from our reflectance learning leads to blurry results and weaker capturing of specularities.}
\label{fig:abl_surf_norm}
\end{figure*}

\subsection{Relighting}
\label{sec:relighting}
Our network is trained on the light stage data with discrete 150 light directions. 
However, it allows us to continously sample light directions at test time, see Sec.~\ref{sec:reflectance_field_repr}.
Given an input image, we can also estimate the scene illumination using OLAT predictions of these 150 light directions. 
Since light transport is additive, the final image under any arbitrary environment map can be written as $\sum_{l=0}^N \lambda_l \hat{I_t}(T_s, N_s^c, N_t^c, \mathbf{\omega_l})$. 
$N$ is the number of light sources, which determines the resolution for the environment. 
A larger value of $N$ allows for representing the illumination at a high resolution, at the cost of computational efficiency since we need a forward pass of the network to compute each $\hat{I_t}$.
The weights $\lambda_l \in \mathbb{R}^3$ are 
color values
of the environment map at the pixel corresponding to light direction $\mathbf{\omega_l}$.%

\paragraph{Light Estimation} We can also estimate the environment map from an in-the-wild image. 
Given our reflectance field, we can optimize for the final reconstruction as follows:
\begin{equation}
\label{eq:leastsq}
    \mathbf{\lambda^*} 
    =
    \argmin\limits_{
    \{\lambda\}
    }\,\Bigl\|
    \sum\nolimits_{l=0}^{N-1} 
    \lambda_l M_t \odot \hat{I_t}(\mathbf{\omega_l}) - M_t \odot I_t \Bigr\|_2^2
    \enspace{.}
\end{equation}

Here, $I_t$ is an in-the-wild image and 
$\{\lambda\} = \{\lambda_i | i\in\{0,\cdots,N{-}1\}\}$.
We minimize this term using least-squares.
In order to get more detailed reconstruction, we further optimize the light using the feature loss as $\mathbf{\lambda^*} = \argmin_{\{\lambda\}} \|V_f(\hat{T_t}(\mathbf{\omega_l})) - V_f(T_t(I_t)) \|_2^2$, where $T_t$ is the texture map computed from the input image $I_t$. 
We use Adadelta solver~\cite{adaDeltaC} to minimize this term and use the solution of Eq.~\ref{eq:leastsq} as the initialization.

\section{Results}
We perform experiments on in-the-wild images from CelebA-HQ~\cite{karras2017progressive} as well as on controlled light stage data with ground truth available. 
Since all images in our training data include an eye-closed expression, we cannot learn the reflectance of open eyes, and we remove this region from results. 
For quantitative evaluations, we use the scale-invariant mean square error (Si-MSE)~\cite{Zhou_2019_ICCV}
and face dissimilarity metric (Face dis). Face dissimilarity is obtained by measuring euclidean distance between features of ground truth and predicted images using a facial recognition tool~\cite{dlib09}. 
 
\subsection{Qualitative Results}

We perform several experiments to qualitatively evaluate our approach.
Fig.~\ref{fig:results1} shows results from the light stage test data (identity not included in training), with the corresponding ground truths. 
We can synthesize different OLATs with different head poses, closely resembling ground-truth. 
We can capture strong shadows, specularities and sub-surface scattering effects. 
Fig.~\ref{fig:results2} additionally shows relighting results on natural images with different environment maps.
Here, we add the results of many light sources.
Our approach can synthesize results with photorealistic pose-dependent illumination effects, as can be seen in results of faces in different poses.
In Fig.~\ref{fig:comp1} we compare our reconstructions with the monocular reconstruction methods of Smith~\etal~\cite{smith2020morphable}, Schneider~\etal~\cite{Schneider17} and Tewari~\etal~\cite{tewari17MoFA}.
These methods also estimate the scene illumination.
Tewari~\etal~assume faces to be diffuse, Smith~\etal~add a specular component, while Schneider~\etal~use precomputed radiance transfer to model shadows with a diffuse surface assumption. 
We train the approach of Tewari~\etal~\cite{tewari17MoFA} on our training data. Thus, it can be considered as a baseline result where the reflectance model is constrained to be diffuse. 
Smith~\etal~\cite{smith2020morphable} and Schneider~\etal~\cite{Schneider17} are analysis-by-synthesis methods.
Our approach clearly produces more photorealistic reconstructions that better capture specularities, subsurface scattering and shadows. 
The comparison with Smith~\etal~specifically shows the advantages of our representation since their model is also trained on a light stage dataset. 
Fig.~\ref{fig:relight} shows further relighting results where the target environment map is computed from another reference image.
Results show that our reflectance is well disentangled from illumination, even under strong directional colored illumination. %
Our results outperform the state of the art both in terms of the quality of reflectance as well as the quality of scene illumination captured. 
All competing approaches use a spherical harmonic light assumption, which would be incapable of handling high-frequency light conditions, which often lead to strong shadows.
Methods such as ~\cite{yamaguchi_high-fidelity_2018,lattas2020avatarme} do not estimate the scene illumination. 
This makes it difficult to objectively compare to these approaches, especially since every method assumes a different coordinate system making it difficult to visualize the results under the same lighting. 

\subsection{Quantitative Evaluations}
We evaluate our approach quantitatively through a number of experiments. %
Tab.~\ref{tab:olat_metric}
summarizes our OLAT reflectance reconstruction results on the light stage data, on a subset of the test set (40 identities, 8 poses).
The input images were synthesized using 160 natural environment maps, see Sec.~\ref{sec:dataset}. 
A total of 3900 input images are reconstructed with a target pose same as in the input, and 8100  images with a different target pose. 
Tab.~\ref{tab:olat_metric} shows that while our approach produces a lower scale invariant MSE (Si-MSE) for results synthesized with the same pose, the errors only slightly increase with a different pose.
Tab.~\ref{tab:recon_metric} compares our monocular reconstruction on in-the-wild images with that of different approaches \cite{smith2020morphable,tewari17MoFA,Schneider17}. 
We use 130 images from CelebA-HQ~\cite{CelebAHQ} as a test set and report the Si-MSE~\cite{Zhou_2019_ICCV} and face identity dissimilarity (Face dis)~\cite{dlib09}. 
While Si-MSE only looks at pixel-level similarities between images, Face dis uses a face recognition network to compute distances between facial identity embeddings. 
Input images were selected to cover a rich variety in terms of pose and illumination.
Our approach significantly outperforms existing approaches as reported by the lower Si-MSE error and Face dis metrics. 
We also evaluate the quality of reflectance under a ``reflectance transfer'' operation. 
Here, we take two images of the same person in different poses and different natural light conditions from the light stage data. 
We reconstruct the reflectance of both images, and then exchange them before evaluating the reconstruction error. 
This evaluation tests the quality of reflectance under different poses and light conditions. 
We also compare to other methods~\cite{smith2020morphable,Schneider17,tewari17MoFA} in the same manner. 
Tab.~\ref{tab:recon_metric} shows that our approach outperforms these methods over 86 images from our test set.

\subsection{Ablative Study}
We evaluate the different components of our method using several ablative studies. 

\subsubsection{Surface normals}
We assess the importance of providing surface normals as input in the network. 
For this we trained a model without providing the source and target surface normals as input to the reflectance network. 
The network in this case would not have access to the face geometry and head pose. 
Tab.~\ref{tab:olat_metric_normal}
summarizes the results of this experiment. Here, we evaluate OLAT reflectance reconstruction on the light stage data, on a subset of the test set (40 identities, 8 poses).
The input images were synthesized using 160 natural environment maps. 
A total of 3900 input images are reconstructed with a target pose same as in the input, and 8100  images with a different target pose.
This is the same test data used in Tab.~1 of the main paper. 
We report scale invariant MSE (Si-MSE) for renderings with same and different input pose. 
Results show that removing normal maps degrades results noticeably, showing that geometry and pose information is important for the task.
This reduction in performance is also reflected visually in Fig.~\ref{fig:abl_surf_norm} where removing surface normals leads to blurry results and weak specularities.

\subsubsection{Impact of accurate geometry}
To assess the importance of accurate geometry in our solution we train a network which only uses the mean template face mesh. 
The geometry network here only predicts the head pose, without the identity and expression geometry parameters.
We use 130 images from CelebA-HQ~\cite{CelebAHQ} as a test set and report the Si-MSE~\cite{Zhou_2019_ICCV} and face identity dissimilarity (Face dis)~\cite{dlib09}. 
This is the same test-set used in Tab.~2 (main paper).
Tab.~\ref{tab:olat_metric_geom} reports the Si-MSE and face identity dissimilarity over the test-set.
Not learning the face geometry and using a fixed mean mesh instead leads to clear degradation in performance in terms of Si-MSE.

\section{Discussion and Limitations}
While we show results which allow for estimation of full reflectance fields from monocular images for the first time, our method still has some limitations. 
As mentioned before, our method cannot estimate the reflectance of open eyes, since the training dataset does not include such images. 
However, our method successfully generalizes to in-the-wild images for the visible regions, even for different expressions.
Our method in general is limited to the face region, because of geometry reconstruction. 
With advances in more complete monocular geometry reconstruction, including hair and body, our method should be able to estimate more complete reflectance fields.
Although our approach can reconstruct all aspects of reflectance, strong effects such as specularities and strong shadow boundaries can still be a bit blurred, see Fig.~\ref{fig:results1}. 
This could again be due to inaccuracies in monocular reconstruction, leading to misalignments between views during training. 
Nevertheless, we believe that our method takes an important step towards learning and rendering the full reflectance field of a face.

\paragraph{Acknowledgments.} 
We thank Tarun Yenamandra and Duarte David for helping us with the comparisons. 
This work was supported by the ERC Consolidator Grant 4DReply (770784). We also acknowledge support from Technicolor and Interdigital.

{\small
\bibliographystyle{ieee_fullname.bst}
\bibliography{ms.bib}
}

\end{document}